\documentclass[letterpaper]{article} 
\usepackage{aaai2026}  
\usepackage{times}  
\usepackage{helvet}  
\usepackage{courier}  
\usepackage[hyphens]{url}  
\usepackage{graphicx} 
\usepackage{natbib}  
\usepackage{caption} 
\usepackage{algorithm}

\usepackage{amsmath}
\usepackage{booktabs}
\usepackage{multirow}
\usepackage{xcolor}
\usepackage{colortbl} 
\usepackage{algpseudocode}

\newcolumntype{L}[1]{>{\raggedright\arraybackslash}p{#1}}  

\usepackage{listings}
\DeclareCaptionStyle{ruled}{labelfont=normalfont, labelsep=colon, strut=off} 
\floatstyle{ruled}
\newfloat{listing}{tb}{lst}{}
\floatname{listing}{Listing}
\title{CliCARE: Grounding Large Language Models in Clinical Guidelines for Decision Support over Longitudinal Cancer Electronic Health Records}
\author {
    Dongchen Li\textsuperscript{\rm 1},
    Jitao Liang\textsuperscript{\rm 1},
    Wei Li\textsuperscript{\rm 2,3}\thanks{Corresponding author.},
    Xiaoyu Wang\textsuperscript{\rm 4},
    Longbing Cao\textsuperscript{\rm 5}\textsuperscript{\rm *},
    Kun Yu\textsuperscript{\rm 6}
}
\affiliations {
    \textsuperscript{\rm 1} College of Computer Science and Engineering, Northeastern University, Shenyang, China\\
    \textsuperscript{\rm 2} National Frontiers Science Center for Industrial Intelligence and Systems Optimization, Shenyang, China\\
    \textsuperscript{\rm 3} Key Laboratory of Intelligent Computing in Medical Image (MIIC), Northeastern University, Shenyang, China\\
    \textsuperscript{\rm 4} Liaoning Cancer Hospital \& Institute, Shenyang, China\\
    \textsuperscript{\rm 5} Macquarie University, Sydney, Australia\\
    \textsuperscript{\rm 6} College of Medicine and Biological Information Engineering, Northeastern University, Shenyang, China\\

    2490254@stu.neu.edu.cn, 2472127@stu.neu.edu.cn, liwei@cse.neu.edu.cn, 
    wangxyz007@hotmail.com, longbing.cao@mq.edu.au, yukun@bmie.neu.edu.cn
}

\begin{document}
\maketitle

\begin{abstract}
Large Language Models (LLMs) hold significant promise for improving clinical decision support and reducing physician burnout by synthesizing complex, longitudinal cancer Electronic Health Records (EHRs). However, their implementation in this critical field faces three primary challenges: the inability to effectively process the extensive length and fragmented nature of patient records for accurate temporal analysis; a heightened risk of clinical hallucination, as conventional grounding techniques such as Retrieval-Augmented Generation (RAG) do not adequately incorporate process-oriented clinical guidelines; and unreliable evaluation metrics that hinder the validation of AI systems in oncology.
To address these issues, we propose CliCARE, a framework for Grounding Large Language Models in \textbf{Cli}nical Guidelines for Decision Support over Longitudinal \textbf{CA}ncer Electronic Health \textbf{RE}cords. The framework operates by transforming unstructured, longitudinal EHRs into patient-specific Temporal Knowledge Graphs (TKGs) to capture long-range dependencies, and then grounding the decision support process by aligning these real-world patient trajectories with a normative guideline knowledge graph. This approach provides oncologists with evidence-grounded decision support by generating a high-fidelity clinical summary and an actionable recommendation.
We validated our framework using large-scale, longitudinal data from a private Chinese cancer dataset and the public English MIMIC-IV dataset. In these settings, CliCARE significantly outperforms baselines, including leading long-context LLMs and Knowledge Graph-enhanced RAG methods. The clinical validity of our results is supported by a robust evaluation protocol, which demonstrates a high correlation with assessments made by oncologists.
\end{abstract}

\begin{links}
    \link{Code}{https://github.com/sakurakawa1/CliCARE}
\end{links}

\section{Introduction}

\begin{figure}[h]
\centering
\includegraphics[width=0.95\linewidth]{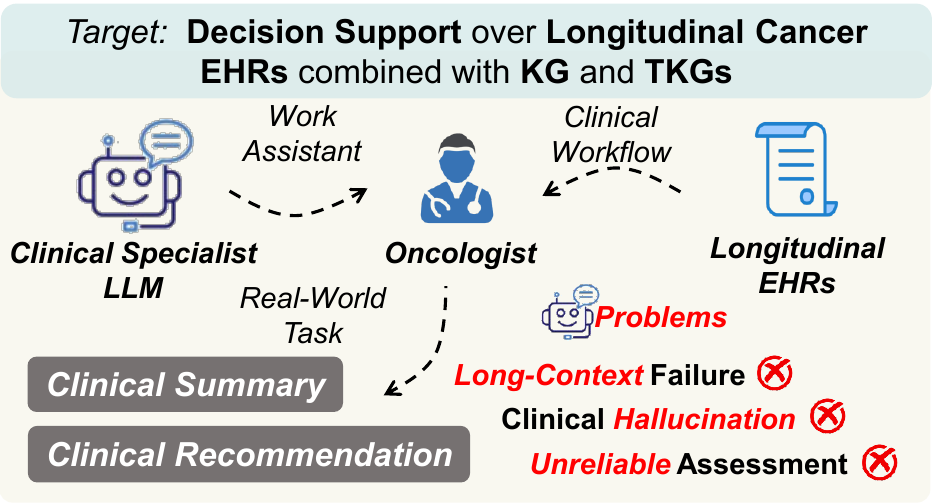}
\caption{The shared challenges for clinicians and LLM in handling complex longitudinal EHRs.}
\label{fig:problem_statement}
\end{figure}

Large Language Models (LLMs) are emerging as promising tools for clinical decision support, with current research demonstrating their potential to serve as collaborative partners that augment expert workflows, alleviate clinician workloads, and improve decision-making in complex fields such as oncology \cite{Hager2024, 10.1145/3613904.3642024}. However, their integration into high-stakes clinical practice is far from straightforward. The reality of clinical oncology involves physicians navigating cognitive burdens from manually integrating fragmented data within multi-year Electronic Health Records (EHRs), a key contributor to professional burnout \cite{warner2020data, sinsky2016allocation}. This chaotic, unstructured environment amplifies the critical disparity between LLMs' high performance on standardized benchmarks and their capabilities in the clinic. Indeed, systematic reviews indicate that their performance in cancer decision-making is inconsistent, with critical safety aspects frequently unaddressed \cite{Hao2025}, while other recent studies confirm that even state-of-the-art models struggle to adhere to treatment guidelines or accurately interpret laboratory results \cite{Hager2024}. Therefore, the frontier of this field is not merely the development of more powerful models, but the creation of robust frameworks that ensure these technologies are reliable, safe, and effectively grounded in expert medical knowledge to augment, not supplant, the role of the physician. In practice, augmenting the expert physician's role means supporting their core clinical workflow: synthesizing a patient's multi-year history into a coherent Clinical Summary, and from that summary, generating an actionable Clinical Recommendation for future treatment.

However, automating this expert workflow with existing LLMs faces three challenges. First, LLMs exhibit an inability to perform effective temporal reasoning over the extensive data records typical of cancer EHRs. Our research addresses a corpus containing large samples of patient records, where a single patient's history can span years, exceed 20,000 tokens, and even include multilingual entries, making brute-force approaches inefficient \cite{liu2024large}. The second challenge is the unacceptable risk of clinical hallucination, which undermines the potential for reliable decision support. Factually incorrect recommendations pose a threat to patient safety, a risk that is exacerbated by the limitations of standard Retrieval-Augmented Generation methods. The retrieval of fragmented text fails to capture the sequential dependencies in a patient's trajectory and cannot effectively bridge the gap with process-oriented clinical guidelines \cite{li2024cancerllm}. Finally, the field confronts two interconnected barriers to real-world adoption. The deployment dilemma centers on a trade-off. On one hand, powerful, closed-source models offer state-of-the-art performance but raise significant concerns regarding cost and patient data privacy. On the other hand, open-source alternatives are more efficient and easier to deploy locally, though they often lag in capability. This trade-off is compounded by the significant challenge of reliable evaluation, as the high-stakes nature of clinical content renders conventional automated metrics untrustworthy, thereby hindering reliable progress and diminishing clinical trust \cite{wang-etal-2023-automated, zheng2023judging}.

To address these barriers, we propose CliCARE, a framework for Grounding Large Language Models in \textbf{Cli}nical Guidelines for Decision Support over Longitudinal \textbf{CA}ncer Electronic Health \textbf{RE}cords. CliCARE first tackles long-context temporal analysis by structuring raw EHRs into Temporal Knowledge Graphs (TKGs) to make temporal relationships explicit (Sec 3.1). It then mitigates hallucinations by grounding the model through a deep alignment of patient trajectories with clinical guidelines (Sec 3.2). This representation provides both guideline-grounded data for fine-tuning specialist models and rich context for large generalist models. Finally, we ensure reliable evaluation via an Expert-Validated LLM-as-a-Judge protocol whose ratings correlate with expert judgments (Sec 4.1).

Our contributions are summarized below:

\begin{itemize}
\item We introduce CliCARE, an end-to-end framework that grounds LLMs by transforming unstructured clinical text from EHRs into TKGs and aligning them with clinical guidelines. It features an adaptable architecture for both generalist and specialist models.
\item We propose a reliable evaluation methodology using an Expert-Validated LLM-as-a-Judge, whose ratings are highly correlated with expert oncologists, addressing the limitations of automated metrics.
\item Extensive experiments on diverse datasets show CliCARE significantly outperforms baselines, while ablation studies confirm the contribution of each component.
\end{itemize}

\begin{figure*}[t!]
    \centering
    \includegraphics[width=1.0\linewidth]{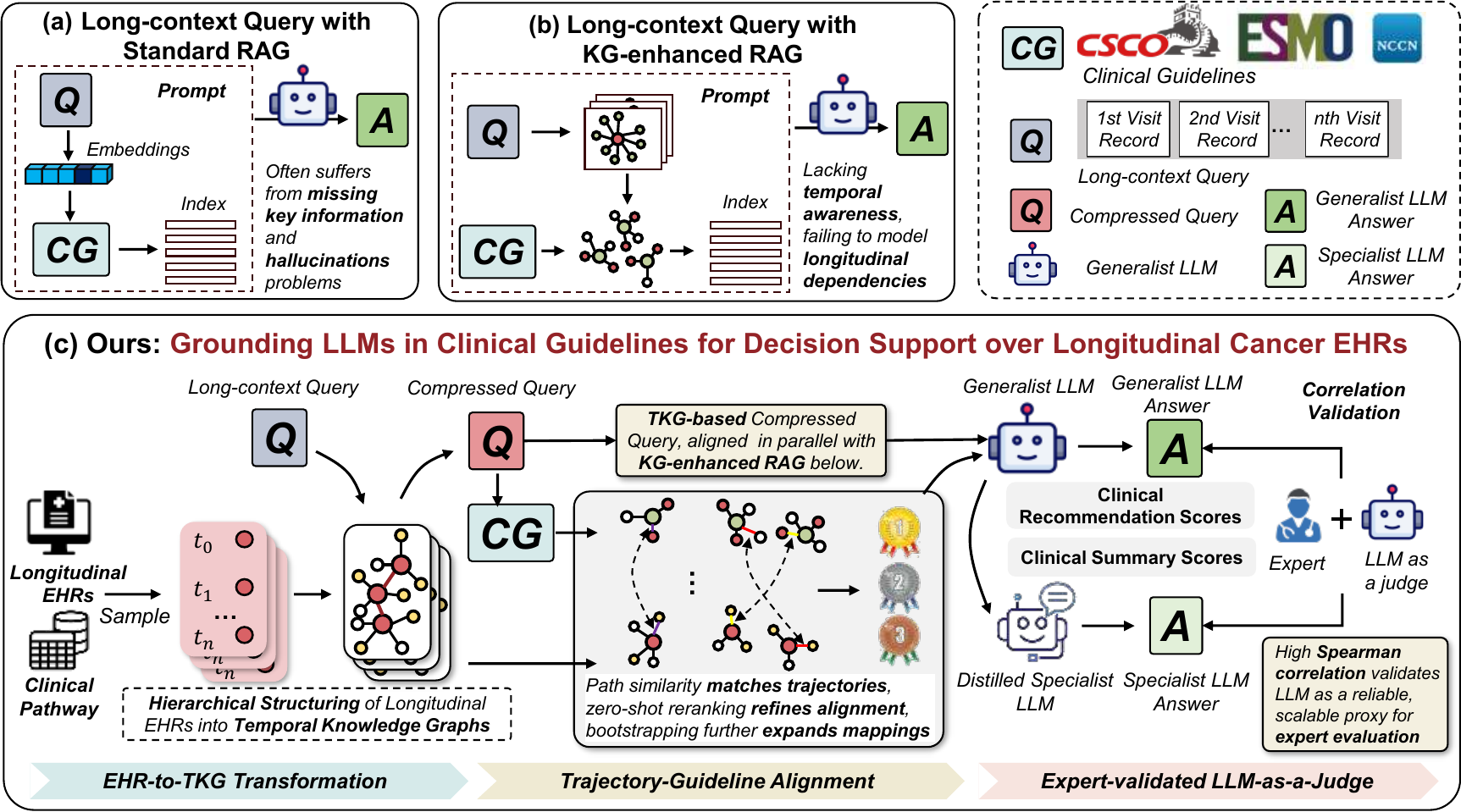}
    \caption{A comparison of RAG approaches for long-form longitudinal clinical tasks. (a) Standard RAG suffers from missing key information and hallucinations. (b) KG-enhanced RAG struggles to model temporal dependencies in patient journeys. (c) In contrast, our CliCARE framework transforms EHRs into Temporal Knowledge Graphs, aligns patient trajectories with guidelines, and generates answers using a distilled specialist model, which are then assessed by our evaluation approach.}
    \label{fig:framework_overview}
\end{figure*}

\section{Related Work}
\subsection{LLMs for Decision Support with Long-Form EHRs}
Leveraging LLMs for clinical decision support holds promise for synthesizing complex EHRs. However, the longitudinal, long-form nature of these records presents a barrier. Models often struggle with long-context processing, leading to issues like the lost-in-the-middle problem and performance degradation \cite{liu2024large, zhang2024leveraging, yang2024fine}. These challenges are now being systematically evaluated by specialized benchmarks like LongBench and MedOdyssey \cite{bai2024longbenchbilingualmultitaskbenchmark, fan2024medodyssey}.

To mitigate these issues and unlock the potential of LLMs for decision support, a primary strategy is to transform unstructured data into structured formats. Current approaches range from prompting LLMs on pre-structured data for prediction tasks \cite{Zhu2024prompting}, or enhancing predictions from structured codes via retrieval augmentation \cite{xu-etal-2024-ram}, to generating Patient Journey Knowledge Graphs (PJKGs) from raw text, though the latter can face reliability challenges \cite{khatib2025patient}. The emergence of sophisticated, multi-agent architectures like ColaCare further raises the bar for comprehensive EHR modeling \cite{wang2025colacare}, highlighting the field's push towards robust, structured solutions for decision support.

\subsection{Knowledge Graph-enhanced LLMs and RAG}
Augmenting LLMs with external Knowledge Graphs (KGs) is a crucial strategy for mitigating factual errors and hallucinations, which is essential for safety in high-stakes domains such as healthcare \cite{khan2024data}. This approach helps bridge the gap between general-purpose models and specialized clinical knowledge \cite{yu2025ragkgil}. However, standard RAG often retrieves isolated text snippets, overlooking the relational structures necessary for clinical decision-making \cite{lewis2020retrieval}. Recent efforts seek to address this by focusing retrieval on specific knowledge sources, such as directly incorporating clinical practice guidelines \cite{oniani2024enhancing} or enhancing multimodal EHR analysis through RAG-driven frameworks like REALM \cite{zhu2024realm}.
To more fundamentally resolve knowledge fragmentation, Graph-Aware RAG moves beyond text retrieval to extract structured information, retrieving entire knowledge paths to improve diagnosis prediction \cite{gao2025leveraging} or pulling relevant subgraphs via frameworks like MedRAG and GNN-RAG \cite{zhao2025medrag, feng2024gnnrag}. An even more advanced frontier moves beyond retrieval to the alignment and fusion of KGs and LLMs at the representation level \cite{jiang2024efficient}, a synergy that creates a virtuous cycle of grounding and enrichment \cite{maushagen2024populating}.

\subsection{Assessment of Open-Ended Clinical Generation Tasks}

Evaluating open-ended generation from LLMs in high-stakes medical domains presents a challenge. Traditional automated metrics, such as ROUGE and BLEU, are regarded as inadequate because their emphasis on lexical overlap fails to capture essential aspects like clinical validity, factual accuracy, and safety \cite{wang-etal-2023-automated, singhal2023large}. In response, research has increasingly focused on more nuanced evaluation methods, including dynamic agent-based assessments \cite{tu2024craft, tadevosyan2024amie} and the scalable LLM-as-a-Judge paradigm \cite{zheng2023judging}. However, the reliability of LLM judges is compromised by known systematic biases like positional bias and verbosity, raising safety concerns in a field where deep domain expertise is essential \cite{zheng2023judging, wang2024large}. This highlights the urgent need for rigorous methodologies to validate automated judgments against human expert reasoning.

Existing research has treated long-context processing, knowledge grounding, and reliable evaluation as distinct challenges. A research gap exists in developing a solution that simultaneously addresses the long-context limitations in real-world EHRs, provides deep grounding in clinical guidelines that exceeds standard RAG, and guarantees trustworthy assessment. CliCARE is designed to bridge this gap by integrating these capabilities into a unified pipeline.


\section{CliCARE}

In this section, we present the \textbf{CliCARE} framework, as illustrated in Figure~\ref{fig:framework_overview}. CliCARE is designed to systematically analyze long-form, unstructured cancer EHRs to generate a clinically grounded clinical summary and clinical recommendations. A key feature of this design is its extensibility: the guideline knowledge graph can be efficiently updated and expanded to accommodate new clinical evidence and corresponding guidelines.

\subsection{EHR-to-TKGs Transformation}
The initial stage of CliCARE transforms raw, multi-year EHRs from unstructured text into patient-centric TKGs, effectively addressing the fundamental challenge of long-context temporal reasoning.
\subsubsection{Event Extraction.}
The complete EHR for each patient $p$ can be formalized as a sequence of documents $D_p=(d_{\tau_1}, d_{\tau_2}, \dots, d_{\tau_n})$ ordered by timestamps $T_p=\{\tau_1, \tau_2, \dots, \tau_n\}$, where each document $d_{\tau_i}$ is an unstructured or semi-structured clinical text at time $\tau_i$. To manage this extensive text sequence, we developed an efficient context processing pipeline, $f_{\text{pipeline}}$, to systematically compress, refine, and structure the raw text before it is input into the final pathway generation model.
\begin{equation}
E_p = f_{\text{pipeline}}(D_p)
\end{equation}
Here, $E_p$ represents a structured sequence of key clinical events. Specifically, the core of $f_{\text{pipeline}}$ is an extractive summarization module based on the Longformer model. Given the computational cost of processing the entire $D_p$, we partition the document sequence into the most recent clinical note, $d_{\tau_n}$, and the historical records, $D_{p}^{\text{hist}} = (d_{\tau_1}, \dots, d_{\tau_{n-1}})$. We utilize a Longformer model \cite{beltagy2020longformer}, $\mathcal{M}_{\text{LF}}$, pre-trained on clinical text, to process the extensive historical records $D_{p}^{\text{hist}}$:
\begin{equation}
S_p^{\text{hist}} = \mathcal{M}_{\text{LF}}(D_{p}^{\text{hist}}; \theta_{\text{LF}})
\end{equation}
where $\theta_{\text{LF}}$ are the model parameters and $S_p^{\text{hist}}$ is a summary text that includes the most informative sentences from the historical records. This summary effectively functions as the patient's past medical history. The most recent clinical note, $d_{\tau_n}$, is regarded as the history of the present illness. The final structured event sequence $E_p$ is created by concatenating these two components, thus providing a chronologically coherent and condensed patient history. From this combined text, key clinical facts—such as diagnostic confirmations, staging updates, treatment regimens, biomarker trends, and imaging assessments—are identified and organized into discrete events using BERT for information extraction\cite{yang2021medbert, huang2019clinicalbert}.

\subsubsection{TKG Instantiation.}
Extracted event sequences $E_p$ are organized into a patient-centric Clinical TKG, denoted as $G_t = (E_t, R_t, T)$, where $E_t$ is the set of entities, $R_t$ the set of relations, and $T$ the set of timestamps. To enrich the TKG with standardized medical knowledge, we first construct a general, static biomedical knowledge graph $G_B = (\mathcal{E}_B, \mathcal{R}_B)$, where $\mathcal{E}_B$ contains standardized medical concepts and $\mathcal{R}_B$ represents the relations. For each patient, letting $\mathcal{E}_p$ be the set of raw clinical entities extracted from the patient's record, we instantiate a personalized TKG $G_t$ by linking extracted clinical entities from the patient's record to the concepts in $G_B$. This uses an entity linking function $\phi : \mathcal{E}_p \rightarrow \mathcal{E}_B$, which maps textual mentions in the EHR to their corresponding canonical entries in the biomedical ontology. Each entity $e \in \mathcal{E}_t$ is represented as $e = (e_B, \tau, A)$, where $e_B \in \mathcal{E}_B$ is the linked standard entity, $\tau \in T$ is the event timestamp, and $A$ is a set of event-specific attributes.

The TKG employs a hierarchical timestamp granularity by assigning precise timestamps $\tau\in T$ only to macro-level clinical encounters, while linking intra-encounter events through relative temporal relations, thereby mirroring the structure of real-world clinical records designed to capture the dynamic evolution of a patient's disease course.

\begin{figure}[h]
    \centering
    \includegraphics[width=0.47\textwidth]{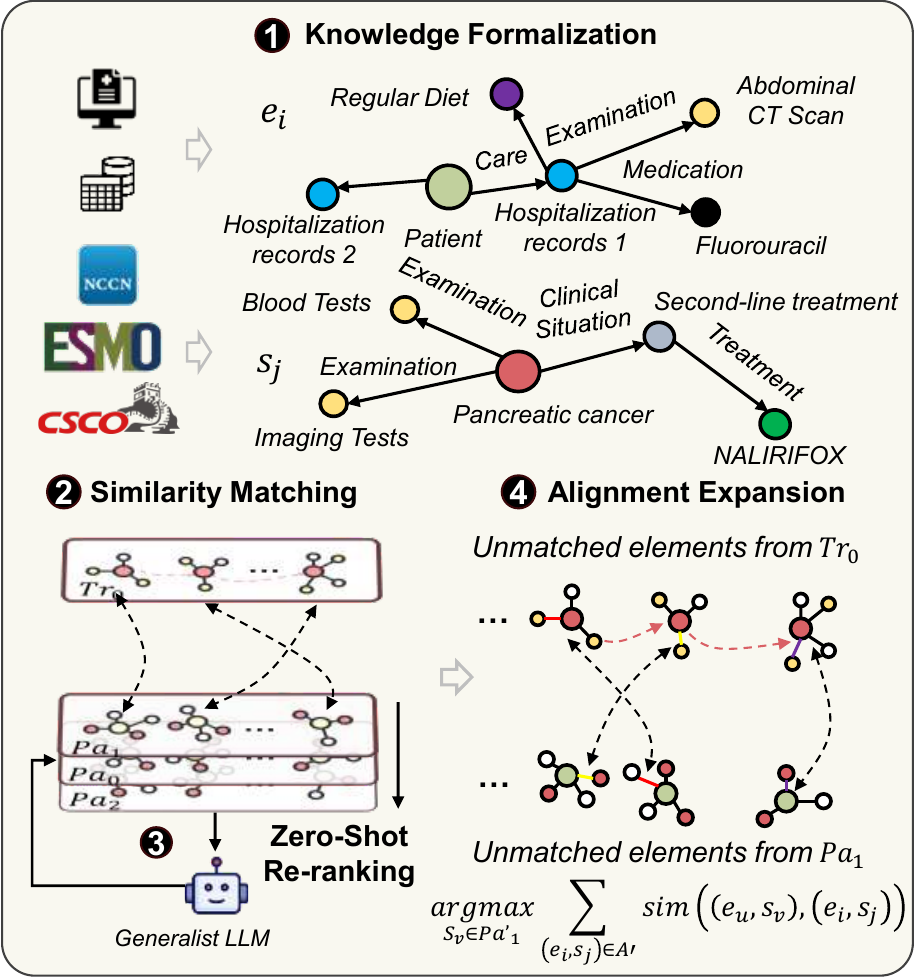}
    \caption{Trajectory-Guideline Alignment workflow. It fuses patient data with guidelines via semantic matching, LLM-based Reranking, and iterative bootstrapping expansion to create a comprehensive, evidence-grounded mapping.}
    \label{fig:alignment_workflow}
\end{figure}

\subsection{Trajectory-Guideline Alignment}
To integrate real-world patient data with normative medical knowledge, this stage aligns the descriptive patient TKG with a prescriptive guideline KG through a training-free fusion pipeline, as illustrated in Figure~\ref{fig:alignment_workflow}.

\subsubsection{Knowledge Formalization.}
Our guideline knowledge graph, $G_g$, is a normative, static graph formalized as $G_g=(E_g, R_g)$. It is constructed based on authoritative CPGs, where $E_g$ represents abstract medical concepts such as \textit{Cancer}, \textit{ClinicalSituation}, and \textit{Treatment}. The edges, $R_g$, represent logical and recommendation relationships, collectively forming a graph that represents an idealized clinical workflow. We first perform a basic entity-level alignment between the patient's temporal graph, $G_t$, and $G_g$ using medical ontologies. Subsequently, the clinical history for each patient $p$ is organized from $G_t$ into a time-ordered sequence of Clinical Events, resulting in a Temporal Trajectory $Tr_p = \langle e_1, e_2, \dots, e_m \rangle$. The set of all patient trajectories is denoted as $\{Tr_p\}_{p=1}^P$. Concurrently, we systematically enumerate all possible normative treatment workflows from $G_g$ to form a set of paths $\{Pa_k\}_{k=1}^K$, where each path $Pa_k = \langle s_1, s_2, \dots, s_l \rangle$ represents a recommended sequence of guideline steps.

\subsubsection{Similarity Matching.}
We developed a global matching strategy based on deep semantic representations to directly assess the similarity between the entire patient trajectory and each candidate guideline path. Specifically, the matching score for a candidate path $Pa_k = \langle s_1, s_2, \dots, s_l \rangle$ with the patient trajectory $Tr_p$ is computed as follows. First, for each step $s_j$ in the guideline path, we identify the most semantically similar event in the patient's event trajectory $Tr_p$. This similarity is calculated using a BERT model, $f_{\text{BERT}}$, pre-trained on biomedical text. Finally, we aggregate the best-match similarities for all steps to derive the total score for the path. Formally, the matching score is defined as:
{\small
\begin{multline}
\text{Score}(Tr_p, Pa_k) = \sum_{j=1}^{l} \max_{e_i \in Tr_p} \Big(
\text{cos\_sim}(f_{\text{BERT}}(\text{desc}(s_j)), \\
f_{\text{BERT}}(\text{desc}(e_i)))
\Big)
\end{multline}}
where the $\text{desc}(\cdot)$ function retrieves the text description of a node, and $\text{cos\_sim}$ computes the cosine similarity between two vectors. A higher score indicates a better alignment between the normative path $Pa_k$ and the patient's experience. The optimal matching path $Pa^*$ is then determined by selecting the candidate with the highest score:
\begin{equation}
Pa^* = \arg\max_{Pa_k} \text{Score}(Tr_p, Pa_k)
\end{equation}
This method relies primarily on the deep semantic understanding provided by BERT, transcending simple lexical matching to capture conceptual associations between clinical events and guideline steps, thereby facilitating precise trajectory-path alignment.

\subsubsection{LLM-based Reranking.}
The method described above generates a ranked list of candidate alignment paths for each patient trajectory. However, purely algorithmic matching may fail to capture the complexities of clinical logic. Therefore, we introduce an LLM as a Clinical Reasoner, $f_{LLM}$, to perform reranking. We provide the LLM with a rich-context prompt in a zero-shot manner, which includes the patient's trajectory $Tr_p$, the top-N candidate normative paths $\{Pa_1, \dots, Pa_N\}$, and their corresponding matching scores $\{\text{Score}_1, \dots, \text{Score}_N\}$. The LLM's task is to evaluate which candidate alignment is the most clinically plausible and to output a reranked list: $\langle Pa'_1, \dots, Pa'_N \rangle = f_{LLM}(Tr_p, \{\langle Pa_k, \text{Score}_k \rangle\}_{k=1}^N)$.

\subsubsection{Alignment Expansion.}
To further enhance the coverage and accuracy of our alignments, we introduce an expansion stage inspired by bootstrapping techniques \cite{sun2018bootstrapping}. After the LLM reranking, the top-ranked alignment path, $Pa'_1$, and its corresponding aligned node pairs serve as a high-confidence seed set, $A'$. We then iteratively expand this set. For each unaligned event $e_u$ in the patient trajectory, the framework seeks to identify the best corresponding node $\hat{s}$ from the entire guideline path $Pa'_1$. The selection process is not based solely on direct similarity; rather, it considers how well the candidate pair $(e_u, s_v)$ coheres with the entire set of existing high-confidence alignments in $A'$. This is accomplished by selecting the guideline node $s_v$ that maximizes the sum of consistency scores with all established pairs in the seed set. The process is formalized as follows:

\begin{equation}
\hat{s} = \arg\max_{s_v \in Pa'_1} \sum_{(e_i, s_j) \in A'} \text{sim}((e_u, s_v), (e_i, s_j))
\end{equation}
where $\text{sim}$ is a function that measures the consistency between a candidate pair $(e_u, s_v)$ and an existing seed pair $(e_i, s_j)$. This function utilizes the semantic representations derived from the language model $f_{\text{BERT}}$ to compute the similarity between the corresponding nodes within the pairs. A high consistency score indicates that the semantic relationship between the patient event $e_u$ and the candidate guideline node $s_v$ is analogous to the established, high-confidence relationship between the seed event $e_i$ and the guideline node $s_j$. This approach enables us to leverage established strong associations to infer new alignment relationships, thereby expanding our alignment set $A'$. After determining the final expanded alignment path, we employ a principled fusion strategy to enrich the guideline knowledge graph $G_g$ with evidence from the patient trajectory $Tr_p$. Ultimately, this alignment process produces a robust, evidence-fused knowledge representation that serves as a direct, high-quality context for an LLM to generate its final clinical summary and clinical recommendation.

\section{Experiments}

\subsection{Evaluation Method}
To assess the quality of generated text, we developed an \textbf{Expert-Validated LLM-as-a-Judge} component. This component assesses two primary tasks: retrospective Clinical Summary, referred to as $T_\textbf{CS}$, and prospective Clinical Recommendation, referred to as $T_\textbf{CR}$. Our methodology employs a concise, four-dimensional rubric, co-designed with senior oncologists, to assess \textit{Factual Accuracy}, \textit{Completeness \& Thoroughness}, \textit{Clinical Soundness}, and \textit{Actionability \& Relevance}. The LLM judge is prompted to assign a score ranging from 1 (poor) to 5 (excellent) for each dimension.

To address the systematic biases of LLM judges, including positional bias, verbosity, and self-enhancement \cite{zheng2023judging, wang2024large}, and after verifying their presence in our specific context, we implement a robust two-part mitigation protocol. First, to ensure rating stability and reduce model bias, we create a judging ensemble composed of three powerful LLMs, such as GPT-4.1 \cite{achiam2023gpt}, Claude 4.0 Sonnet \cite{anthropic_claude4_2025}, and Gemini 2.5 Pro \cite{comanici2025gemini}, using their averaged score. Second, to eliminate ordering effects, all items are presented in a randomly shuffled sequence during the evaluation.

We validated the ratings of our LLM judge against those of three experienced oncologists using a subset of data to ensure reliability. We employed Spearman's rank correlation coefficient, denoted as $\rho$, a non-parametric measure that assesses the monotonic relationship between the LLM's and the experts' rankings. The coefficient is calculated as:
\begin{equation}
\rho = 1 - \frac{6 \sum d_i^2}{n(n^2 - 1)}
\end{equation}
where $n$ is the number of samples and $d_i$ is the difference in ranks for each sample. A high correlation provides evidence that our LLM judge functions as a reliable and scalable proxy for expert assessment. This justifies its application for large-scale evaluations throughout our experiments.

\subsection{Datasets}
We evaluated our framework using two large-scale clinical datasets. The first dataset is a private Chinese collection, referred to as CancerEHR. It contains longitudinal records for 2,000 patients from Liaoning Cancer Hospital. These records span extensive periods—some exceeding two decades—resulting in inputs of up to 20,000 tokens. The dataset includes a variety of data types, such as physicians' orders, laboratory results, and surgical notes.
The second dataset is derived from the publicly available MIMIC-IV dataset \cite{Johnson2023MIMIC}, filtered to include only patients with cancer-related diagnoses, which we refer to as MIMIC-Cancer. This dataset provides a focus on disease progression similar to CancerEHR; however, the language and data structure differ, offering a robust test of our method's generalizability.
For brevity, in the subsequent implementation and results sections, we will refer to the two datasets as $D_\textbf{CEHR}$ for CancerEHR and $D_\textbf{MC}$ for MIMIC-Cancer. Similarly, the two primary tasks will be abbreviated as $T_\textbf{CS}$ for Clinical Summary and $T_\textbf{CR}$ for Clinical Recommendation.

\subsection{Baselines}
We compare our proposed CliCARE framework against a variety of robust baseline methods. These include standard RAG pipelines implemented with powerful open-source models such as Mistral-7B and its instruction-tuned variant \cite{jiang2023mistral}, Qwen3-8B \cite{qwen3_technical_report_2025}, and the domain-specific BioMistral-7B \cite{labrak2024biomistral}. Additionally, we evaluate more advanced KG-enhanced RAG techniques designed for long-context or knowledge-intensive tasks. The selected methods include BriefContext \cite{zhang2024leveraging}, which employs a Map-Reduce strategy, as well as several Graph-Aware RAG frameworks, such as GNN-RAG \cite{feng2024gnnrag}, KG2RAG \cite{shi2024kg2rag}, and the healthcare-focused MedRAG \cite{zhao2025medrag}. We evaluate these RAG frameworks using both the open-source models and leading generalist models, including Deepseek-R1 \cite{guo2025deepseek}, Gemini 2.5 Pro, GPT-4.1, and Claude-4.0-Sonnet.

\subsection{Implementation Details}
In the knowledge graph alignment stage, the threshold is set to 0.7 when using BERT to calculate semantic cosine similarity in the initial step. During the fine-tuning stage, we divided the 2,000-sample dataset into a training set of 1,800 samples and a test set of 200 samples, with 10\% of the training data reserved for validation. The key hyperparameters for training include a batch size of 1, a maximum context length of 20,000 tokens, and an initial learning rate of $5 \times 10^{-5}$ with a cosine scheduler. We utilized BF16 for mixed-precision training, set the maximum output length to 4,096 tokens, and trained for 3 epochs. All experiments were conducted using a configuration of 4 NVIDIA A800 GPUs.

\begin{table}[t]
    \centering
    \small
    \begin{tabular*}{\columnwidth}{@{\extracolsep{\fill}}lcccc@{}}
    \toprule
    \multirow{2}{*}{\textbf{Method}} & \multicolumn{2}{c}{\textbf{$D_\textbf{CEHR}$}} & \multicolumn{2}{c}{\textbf{$D_\textbf{MC}$}} \\
    \cmidrule(lr){2-3} \cmidrule(lr){4-5}
    & $T_\textbf{CS}$ & $T_\textbf{CR}$ & $T_\textbf{CS}$ & $T_\textbf{CR}$ \\
    \midrule
    \textit{Qwen-3-8B} \\
    \quad StandardRAG        & 1.485 & 1.527 & 2.475 & 2.467 \\
    \quad BriefContext       & \underline{2.681} & \underline{2.701} & \underline{2.571} & \underline{2.497} \\
    \quad MedRAG* & 2.333 & 2.366 & 2.495 & 2.462 \\
    \quad KG2RAG* & 2.595 & 2.558 & 2.317 & 2.166 \\
    \quad GNN-RAG* & 2.508 & 2.527 & 2.194 & 2.182 \\
    \quad \textbf{CliCARE}   & \textbf{3.173} & \textbf{3.215} & \textbf{2.575} & \textbf{2.544} \\
    \midrule
    \textit{Gemini 2.5 Pro} \\
    \quad StandardRAG        & 2.735 & 2.818 & 3.563 & 3.556 \\
    \quad BriefContext       & \underline{4.527} & 4.468 & 4.354 & 4.233 \\
    \quad MedRAG* & 4.470 & \underline{4.576} & \textbf{4.476} & \underline{4.323} \\
    \quad KG2RAG* & 3.845 & 3.942 & 3.747 & 3.797 \\
    \quad GNN-RAG* & 3.607 & 3.552 & 3.683 & 3.588 \\
    \quad \textbf{CliCARE}   & \textbf{4.976} & \textbf{4.965} & \underline{4.398} & \textbf{4.333} \\
    \bottomrule
    \end{tabular*}
    \caption{
    CliCARE Outperforms RAG Baselines on Clinical Generation Tasks. Scores are assigned by our Expert-Validated LLM-as-a-Judge. The asterisk (*) denotes KG-enhanced RAG variants.
    }
    \label{tab:main_results}
\end{table}

\begin{table*}[t]
    \centering
    \small
    \begin{tabular*}{\textwidth}{@{\extracolsep{\fill}}lcccccccc@{}}
    \toprule
    \multirow{3}{*}{\textbf{Method}} & \multicolumn{4}{c}{\textbf{Standard RAG}} & \multicolumn{4}{c}{\textbf{CliCARE}} \\ \cmidrule(lr){2-5} \cmidrule(lr){6-9} 
     & \multicolumn{2}{c}{\textbf{$D_\textbf{CEHR}$}} & \multicolumn{2}{c}{\textbf{$D_\textbf{MC}$}} & \multicolumn{2}{c}{\textbf{$D_\textbf{CEHR}$}} & \multicolumn{2}{c}{\textbf{$D_\textbf{MC}$}} \\ \cmidrule(lr){2-3} \cmidrule(lr){4-5} \cmidrule(lr){6-7} \cmidrule(lr){8-9}
     & $T_\textbf{CS}$ & $T_\textbf{CR}$ & $T_\textbf{CS}$ & $T_\textbf{CR}$ & $T_\textbf{CS}$ & $T_\textbf{CR}$ & $T_\textbf{CS}$ & $T_\textbf{CR}$ \\ \midrule
    Mistral-v0.1-7B & 1.120 & 1.164 & 2.505 & 2.505 & 1.407\,(+0.287) & 1.526\,(+0.362) & 2.575\,(+0.070) & 2.514\,(+0.009) \\
    Mistral-Instruct-v0.1-7B & 1.054 & 1.070 & 2.183 & 2.115 & 1.274\,(+0.220) & 1.355\,(+0.285) & 2.231\,(+0.048) & 2.071\,(-0.044) \\
    Biomistral-7B & 1.161 & 1.098 & \textbf{2.785} & \textbf{2.698} & 1.548\,(+0.387) & 1.529\,(+0.431) & \textbf{2.903}\,(\textbf{+0.118}) & \textbf{2.742}\,(+0.044) \\
    Qwen-3-8B & \textbf{1.485} & \textbf{1.527} & 2.475 & 2.467 & \textbf{3.173}\,(\textbf{+1.688}) & \textbf{3.215}\,(\textbf{+1.688}) & 2.575\,(+0.100) & 2.544\,(\textbf{+0.077}) \\
    \midrule
    Gemini-2.5-Pro & \textbf{2.735} & 2.818 & 3.563 & 3.556 & \textbf{4.976}\,(+2.241) & \textbf{4.965}\,(+2.147) & 4.398\,(\textbf{+0.835}) & 4.333\,(\textbf{+0.777}) \\
    GPT-4.1 & 2.667 & 2.873 & \textbf{4.419} & \textbf{4.429} & 4.690\,(+2.023) & 4.703\,(+1.830) & \textbf{4.737}\,(+0.318) & \textbf{4.676}\,(+0.247) \\
    Deepseek-R1 & 2.667 & \textbf{2.878} & 4.016 & 4.000 & 4.946\,(\textbf{+2.279}) & 4.935\,(\textbf{+2.057}) & 4.409\,(+0.393) & 4.319\,(+0.319) \\
    Claude-4.0-Sonnet & 2.417 & 2.624 & 3.898 & 3.868 & 3.893\,(+1.476) & 3.924\,(+1.300) & 4.183\,(+0.285) & 4.110\,(+0.242) \\ \bottomrule
    \end{tabular*}
    \caption{Model performance with standard RAG versus the CliCARE framework. Applying CliCARE provides a substantial performance uplift for most models.}
    \label{tab:comparison_with_without_kg}
\end{table*}

\subsection{Experimental Results}

\subsubsection{High Agreement with Clinician Judgments.}
Acknowledging the limitations of traditional metrics for clinical tasks, we validated our LLM-as-a-Judge protocol against three experienced oncologists. To ensure a feasible yet representative assessment, we created a validation subset by randomly sampling generated outputs from eight different models. Our protocol minimized bias by evaluating these outputs column-wise and presenting the Clinical Summary and Recommendation tasks together for a comprehensive review. The results demonstrate a strong positive correlation between the automated ratings and those of the experts. Specifically, the Spearman's rank correlation $\rho$ between our LLM judge's scores and the physicians' mean scores was approximately 0.7, confirming that our metric serves as a reliable proxy for human expert judgment.

\subsubsection{CliCARE Significantly Outperforms Baselines.}
As detailed in Table~\ref{tab:main_results}, CliCARE demonstrates a clear performance advantage over a suite of robust baselines, including both context-aware and KG-enhanced RAG methods. The benefits of our framework are most pronounced when paired with a powerful model on complex datasets. With Gemini 2.5 Pro, CliCARE achieves impressive Clinical Summary and Recommendation scores of 4.976 and 4.965, respectively, on the challenging CancerEHR dataset. This performance significantly surpasses that of other methods. For instance, while BriefContext achieves a commendable score of 4.527, it does so through a costly Map-Reduce strategy that involves multiple LLM calls, underscoring the efficiency of CliCARE's approach. Even when utilizing a smaller model like Qwen-3-8B, CliCARE obtains scores of 3.173 and 3.215, substantially outperforming all baselines on the same complex dataset. This success is attributed to CliCARE’s TKG transformation, which effectively organizes the chaotic, longitudinal patient records and overcomes the fragmented retrievals that hinder other RAG pipelines.

\subsubsection{Structured Knowledge is Key for Complex EHRs.}
As demonstrated in Table~\ref{tab:comparison_with_without_kg}, our framework's knowledge structuring offers a significant advantage over standard RAG. The performance uplift is most pronounced on the complex CancerEHR dataset, where nearly all models exhibit substantial gains. Notably, the improvements for Qwen-3-8B and Deepseek-R1 are the largest in their respective groups, with their Clinical Summary scores increasing by +1.688 and +2.279, respectively. This underscores that even advanced models require a coherent structure for effective reasoning on complex records. On the simpler MIMIC-Cancer dataset, while the absolute gains are smaller, CliCARE still delivers a distinct and consistent advantage. For instance, it elevates the score of a strong baseline like GPT-4.1 from 4.419 to 4.737, a gain of +0.318. While the uplift is nearly universal, we do note a single case of minor performance degradation, confirming the intricate nature of these tasks.

\begin{table}[h]
    \centering
    \small
    
    \begin{tabular*}{\columnwidth}{@{\extracolsep{\fill}}lcccc@{}}
        \toprule
        \multirow{2}{*}{\textbf{Method}} & \multicolumn{2}{c}{\textbf{$D_\textbf{CEHR}$}} & \multicolumn{2}{c}{\textbf{$D_\textbf{MC}$}} \\
        \cmidrule(lr){2-3} \cmidrule(lr){4-5}
        & $T_\textbf{CS}$ & $T_\textbf{CR}$ & $T_\textbf{CS}$ & $T_\textbf{CR}$ \\
        \midrule

        \textbf{CliCARE (Q)} & \textbf{3.173} & \textbf{3.215} & \textbf{2.575} & \textbf{2.544} \\
        \quad w/o Exp.   & 3.012\,(-) & 3.035\,(-) & 2.075\,(-) & 2.110\,(-) \\
        \quad w/o Rerank & 2.857\,(-) & 2.866\,(-) & 2.000\,(-) & 1.962\,(-) \\
        \quad w/o Comp.  & 1.485\,(-) & 1.527\,(-) & 2.475\,(+) & 2.467\,(+) \\
        \midrule

        \textbf{CliCARE (G)} & \textbf{4.976} & \textbf{4.965} & \textbf{4.398} & \textbf{4.333} \\
        \quad w/o Exp.   & 4.619\,(-) & 4.630\,(-) & 3.737\,(-) & 3.786\,(-) \\
        \quad w/o Rerank & 4.542\,(-) & 4.628\,(-) & 3.774\,(+) & 3.824\,(+) \\
        \quad w/o Comp.  & 2.735\,(-) & 2.818\,(-) & 3.563\,(-) & 3.556\,(-) \\
        \bottomrule 
    \end{tabular*}
    \caption{Ablation study on CliCARE framework components. Q denotes Qwen-3-8B and G denotes Gemini-2.5-Pro. Exp., Rerank and Comp. signify the removal of Alignment Expansion, LLM-based Reranking, and TKG-based Compression, respectively. The symbols (+)/(-) indicate a performance increase/decrease compared to the row above.}
    \label{tab:summary_ablation}
\end{table}

\subsubsection{Ablation Study.}
Our ablation study, with results in Table~\ref{tab:summary_ablation}, reveals the nuanced role of each module. This is most evident for the Qwen model on the simpler MIMIC-Cancer dataset; removing TKG-based Compression paradoxically boosts the scores to 2.475 and 2.467. This result is substantially better than when LLM-based Reranking is removed, which causes a drop to 2.000, suggesting aggressive compression can be counterproductive for shorter records. A similar, though less pronounced, positive effect is observed for the Gemini model under the same conditions. Conversely, on the complex CancerEHR dataset, the consistent, significant performance drops from any ablation highlight that the full, integrated CliCARE framework is crucial for achieving optimal performance.

\begin{table}[h]
    \centering
    \small  
    \setlength{\tabcolsep}{4pt}  
    \begin{tabular*}{\columnwidth}{@{\extracolsep{\fill}}L{3.5cm}cccc@{}}
        \toprule
        \multirow{2}{*}{\textbf{Method}} & \multicolumn{2}{c}{\textbf{$D_\textbf{CEHR}$}} & \multicolumn{2}{c}{\textbf{$D_\textbf{MC}$}} \\
        \cmidrule(lr){2-3} \cmidrule(lr){4-5}
        & $T_\textbf{CS}$ & $T_\textbf{CR}$ & $T_\textbf{CS}$ & $T_\textbf{CR}$ \\
        \midrule
        \textit{CliCARE(Qwen-3-8B)} \\
        \quad All \hspace*{1.9em}(100\%)           & 3.173 & 3.215 & 2.575 & 2.544 \\
        \quad Short \hspace*{1.0em}(0\textasciitilde33\%)    & 3.228 & \textbf{3.345} & \textbf{2.850} & \textbf{2.645} \\
        \quad Medium\hspace*{0.1em}(33\%\textasciitilde66\%)   & \textbf{3.267} & 3.283 & 2.429 & 2.450 \\
        \quad Long \hspace*{1.1em}(66\%\textasciitilde100\%) & 2.976 & 2.983 & 2.460 & 2.533 \\
        \midrule
        \textit{CliCARE(Gemini-2.5-Pro)} \\
        \quad All \hspace*{1.9em}(100\%)           & 4.976 & 4.965 & 4.398 & 4.333 \\
        \quad Short \hspace*{1.0em}(0\textasciitilde33\%)    & 4.982 & 4.937 & 4.362 & 4.311 \\
        \quad Medium\hspace*{0.1em}(33\%\textasciitilde66\%)   & 4.962 & 4.976 & 4.365 & 4.317 \\
        \quad Long \hspace*{1.1em}(66\%\textasciitilde100\%) & \textbf{4.988} & \textbf{4.982} & \textbf{4.467} & \textbf{4.373} \\
        \bottomrule
    \end{tabular*}
    \caption{Performance analysis by EHR length. Segments are stratified by percentile (0-33\%, 33-66\%, 66-100\%). Average token counts for CancerEHR segments are 4875, 6303, 9411; for MIMIC-Cancer, 4070, 5176, 6463.}
    \label{tab:length_ablation}
\end{table}

\subsubsection{Performance Analysis Based on EHR Length.}
Further analysis of record length reveals distinct performance patterns, as detailed in Table~\ref{tab:length_ablation}. The smaller model, Qwen-3-8B, performs optimally on Short length records but exhibits a decline in quality when processing the longest records, particularly with the complex CancerEHR data. In contrast, the more powerful Gemini-2.5-Pro model demonstrates strong and consistent performance across all record lengths. Notably, when guided by the CliCARE framework, it achieves its highest scores on the longest record segments for both datasets. This suggests that CliCARE effectively organizes extensive clinical histories and enables advanced models to leverage richer context for enhanced reasoning.

\section{Conclusion}
We introduced CliCARE, a framework for reliable clinical decision support that transforms cancer EHRs into Temporal Knowledge Graphs and aligns them with clinical guidelines. This approach addresses key challenges in long-context reasoning and hallucination, enabling both small specialist and large generalist models to significantly outperform baselines. We also validated a robust LLM-as-a-Judge protocol that correlates highly with expert oncologist assessments, representing a significant advancement toward deploying trustworthy AI in clinical practice. Future work should focus on generalizing this framework to a wider range of clinical domains.

\section*{Acknowledgments}
This work was supported by the Natural Science Foundation of Liaoning Province, China (No. 2025-MS-036) and the Doctoral Student Scientific Research Innovation Project of Northeastern University (No. 02190022125036).
\bibliography{aaai2026}

\clearpage
\appendix

\section{Human Evaluation Questionnaire and Protocol}
\begin{table*}[htbp]
\centering
\caption{Evaluation Rubric for Factual Accuracy, Completeness \& Thoroughness, Clinical Soundness, and Actionability \& Relevance.}
\label{tab:dx_eval_rubric_english}
\resizebox{\textwidth}{!}{%
\begin{tabular}{@{}l@{}}
\toprule
\textbf{Evaluation Dimension \& Scoring Criteria} \\
\midrule
\textbf{Factual Accuracy} \\
\multicolumn{1}{p{0.85\textwidth}}{%
- 5: All key information is 100\% accurate. \newline
- 3: Contains errors in non-critical information or factual deviations that \textbf{do not affect final treatment decisions or patient safety}. \newline
- 1: Contains any \textbf{major factual error that could affect treatment decisions or patient safety}.} \\
\midrule
\textbf{Completeness \& Thoroughness} \\
\multicolumn{1}{p{0.85\textwidth}}{%
- 5: Perfectly covers all critical aspects of the patient's situation, identifies all key data elements, and insightfully adds important potential risks. \newline
- 3: Covers most core aspects and data elements but omits some minor details or has individual improper handling of data. \newline
- 1: Seriously lacks core content, or seriously omits or misunderstands key core data elements.} \\
\midrule
\textbf{Clinical Soundness} \\
\multicolumn{1}{p{0.85\textwidth}}{%
- 5: All conclusions and recommendations are robust, safe, and reflect the clinical prudence of a senior expert. They are implicitly or explicitly based on recognized clinical guidelines. \newline
- 3: Core recommendations are reasonable, but may include some unimportant or slightly unusual minor suggestions, or some recommendations lack a clear evidence-based foundation. \newline
- 1: Contains any recommendations that could jeopardize patient safety, clearly violate clinical common sense, or are based on misquotes.} \\
\midrule
\textbf{Actionability \& Relevance} \\
\multicolumn{1}{p{0.85\textwidth}}{%
- 5: Provides highly insightful, quantifiable, and personalized action plans that focus on solving the most urgent current problems. \newline
- 3: Offers some actionable advice, but some key parts are too general, or recommendations are mixed with retrospective analysis not directly relevant to the immediate next steps. \newline
- 1: Provides a list of invalid information with no guiding value, or the recommendations are entirely disconnected from the current core clinical problem.} \\
\bottomrule
\end{tabular}%
}
\end{table*}

Using a custom-built online questionnaire platform, we instructed practicing oncologists to evaluate two generated outputs: the Clinical Summary and the Clinical Recommendation.

For each evaluation task, the clinician was presented with the following materials:

\begin{itemize}
    \item A patient's complete longitudinal Electronic Health Record (EHR) from the CancerEHR dataset, containing multiple encounters (\texttt{record\_1}, \texttt{record\_2}, etc.).
    \item A human-expert-authored, gold-standard Clinical Summary and a corresponding Clinical Recommendation (collectively, the "label"), which together provide the ideal summary and recommendation based on the EHR.
    \item Anonymized outputs from a random sample of eight models—including both locally deployed (\texttt{CliCARE\_8B},etc.) and API-based (\texttt{CliCARE},etc.) systems—were compared against each other and the gold-standard labels.
\end{itemize}

The clinicians were then instructed to provide the following series of assessments:

\begin{itemize}
    \item Overall Head-to-Head Comparison: The overall quality of the 8 models was directly compared using a 5-point Likert scale, ranging from 1 ("Very Poor") to 5 ("Very Good").
    \item Task-Specific Head-to-Head Comparison: Separate head-to-head comparisons were performed for two key tasks: "Clinical Summary" and "Clinical Recommendations." This evaluation focused specifically on the quality of these two components within the generated output.
\end{itemize}
\section{CancerEHRs Dataset Details and Demographics}
The CancerEHR dataset is a unique, non-public collection of Chinese Electronic Health Records sourced from a large, specialized cancer hospital in China. The dataset underwent a carefully designed, multi-step processing pipeline to ultimately create formatted text suitable for LLM input.

\subsection{Data Processing Pipeline.}
In practice, the raw EHR data underwent a series of processing steps. The core objective was to consolidate the heterogeneous and fragmented raw data from the Hospital Information System (HIS) into a patient-centric, chronologically organized, unified text format. We designed and wrote a series of specialized Python parsing scripts for different clinical data tables. For instance, \texttt{12\_CDR\_OUTPATIENT\_ORDER.py} processed outpatient medical orders, \texttt{3\_CDR\_PATIENT\_DETAIL.py} extracted basic patient information, and other scripts handled inpatient records, lab reports, and imaging descriptions. These scripts accurately extracted key fields from their respective CSV source files. After parsing, each script converted the structured CSV information into semi-structured TXT files. All text fragments extracted from different sources were ultimately sorted and aggregated by patient ID and timestamp. Through this process, we generated a longitudinal text file named \texttt{patient\_inpatient.txt} for each patient in the database. This file comprehensively documents all relevant clinical events for the patient, transforming the scattered, structured data into a patient-centric, sequential, unstructured text, which provides a high-quality input for the summarization and knowledge graph construction in Stage 1.

\subsection{Demographics and Clinical Characteristics.}

We randomly sampled the complete records of 2,000 patients from the CancerEHR dataset for our analysis. Table~\ref{tab:cancer-ehr-distribution} and ~\ref{tab:cancer-ehr-length-stats} displays the demographic details and the distribution of major cancer types within this dataset.

\begin{table}[ht]
    \small
    \centering
    \caption{Top 10 Cancer Type Distribution in the CancerEHRs Dataset.}
    \label{tab:cancer-ehr-distribution}
    \begin{tabular}{lr}
    \toprule
    \textbf{Diagnosis Category} & \textbf{Count} \\
    \midrule
    Breast Cancer & 491 \\
    Malignant Tumor & 271 \\
    Rectal Cancer & 144 \\
    Lung Cancer & 108 \\
    Gastric Cancer & 93 \\
    Colon Cancer & 85 \\
    Ovarian Cancer & 57 \\
    Cervical Cancer & 55 \\
    Lymphoma & 33 \\
    Postoperative & 32 \\
    \bottomrule
    \end{tabular}
\end{table}

Table~\ref{tab:cancer-ehr-distribution} shows the top 10 cancer type distribution in the CancerEHRs dataset.

\begin{table}[ht]
    \centering
    \small
    \caption{Text Length Statistics for Records in the CancerEHRs Dataset.}
    \label{tab:cancer-ehr-length-stats}
    \resizebox{\columnwidth}{!}{
    \begin{tabular}{lrrr}
    \toprule
    \textbf{Statistic} & \textbf{Word Count} & \textbf{Character Count} & \textbf{Digit Count} \\
    \midrule
    Mean & 5,698.96 & 10,103.73 & 799.37 \\
    Median & 5,158.50 & 8,797.50 & 492.50 \\
    Maximum & 21,915.00 & 42,154.00 & 6,803.00 \\
    \bottomrule
    \end{tabular}
    }
\end{table}

Table~\ref{tab:cancer-ehr-length-stats} summarizes the text length statistics for the CancerEHRs dataset.

\section{Processed MIMIC-Cancer Dataset Details and Demographics}
To validate the generalizability of our model across different languages and data sources, we constructed the Processed MIMIC-Cancer Dataset. The processing pipeline for this dataset is similar to that of CancerEHR, but its source is the public MIMIC-IV database.

\subsection{Data Processing Pipeline.}
We first filtered the MIMIC-IV database to select all patients who had an ICD-9 or ICD-10 cancer diagnosis code in their \texttt{diagnoses\_icd.csv} file. Subsequently, we removed admission events from these patients' records that were not directly related to cancer to ensure that each record focuses on the cancer diagnosis and treatment process. Next, we applied a processing pipeline similar to the one described in Section B. The main difference was that we wrote parsing scripts adapted to the MIMIC-IV data schema, extracted data from files such as \texttt{admissions.csv}, \texttt{chartevents.csv}, and \texttt{labevents.csv}, and integrated this information into patient-centric clinical narratives in English. During the KG construction phase for this dataset, we primarily relied on international guidelines such as NCCN and ESMO, as they are more relevant to clinical practice in the United States. The entity recognition and linking processes were also adapted for English medical terminology.
Ultimately, we obtained formatted treatment trajectory texts for 2,000 patients, which could be directly used for comparative experiments with the CancerEHR dataset.

\subsection{Demographics and Clinical Characteristics.}

Table~\ref{tab: MIMIC-IV-distribution} and ~\ref{tab: MIMIC-IV-length-stats} presents the demographic and clinical details for the 2,000 cancer patients selected for our Processed MIMIC-Cancer Dataset.

\begin{table}[ht]
    \centering
    \small
    \caption{Top 10 Cancer Type Distribution in the Processed MIMIC-Cancer Dataset.}
    \label{tab: MIMIC-IV-distribution}
    \begin{tabular}{lr}
    \toprule
    \textbf{Diagnosis Category} & \textbf{Count} \\
    \midrule
    Diffuse large b & 96 \\
    Multiple myeloma & 86 \\
    Acute myeloid leukemia & 66 \\
    Acute myeloblastic leukemia & 53 \\
    B & 52 \\
    Non & 44 \\
    Malignant neoplasm of bronchus & 43 \\
    Liver cell carcinoma & 37 \\
    Malignant neoplasm of prostate & 36 \\
    \bottomrule
    \end{tabular}
\end{table}

Table~\ref{tab: MIMIC-IV-distribution} shows the top 10 cancer type distribution in the processed MIMIC-Cancer Dataset.

\begin{table}[ht]
    \centering
    \small
    \caption{Text Length Statistics for Records in the Processed MIMIC-IV Dataset.}
    \label{tab: MIMIC-IV-length-stats}
    \resizebox{\columnwidth}{!}{
    \begin{tabular}{lrrr}
    \toprule
    \textbf{Statistic} & \textbf{Word Count} & \textbf{Character Count} & \textbf{Digit Count} \\
    \midrule
    Mean & 3,377.30 & 22,895.11 & 1,274.10 \\
    Median & 1,646.00 & 11,538.50 & 526.50 \\
    Maximum & 31,713.00 & 204,741.00 & 16,250.00 \\
    \bottomrule
    \end{tabular}
    }
\end{table}

Table~\ref{tab: MIMIC-IV-length-stats} summarizes the text length statistics for the processed MIMIC-Cancer Dataset.

\begin{figure*}[ht]
    \centering
    \includegraphics[width=0.9\linewidth]{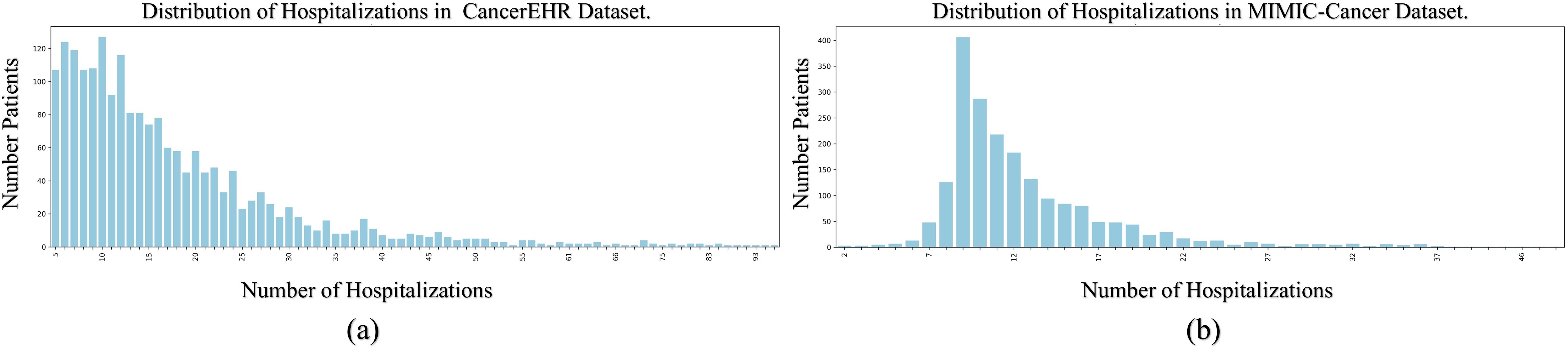}
    \caption{Distribution of Hospitalizations in the CancerEHR Dataset(a) and MIMIC-Cancer Dataset(b).}
    \label{fig:hosp-dist}
\end{figure*}

\begin{figure*}[ht]
    \centering
    \includegraphics[width=0.9\linewidth]{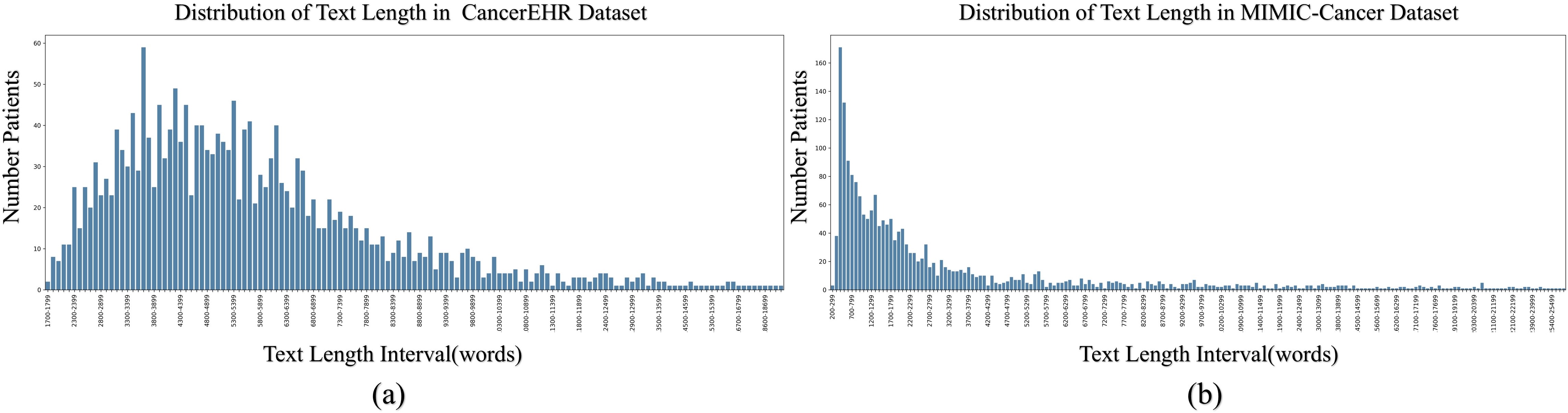}
    \caption{Distribution of Text Length in the CancerEHR Dataset(a) and MIMIC-Cancer Dataset(b).}
    \label{fig:textlen-dist}
\end{figure*}

As shown in Figure~\ref{fig:hosp-dist} and Figure~\ref{fig:textlen-dist}, we present the distribution of the number of hospitalizations and the distribution of text length for patients in both the CancerEHR and processed MIMIC-Cancer Datasets. The hospitalization distribution illustrates the number of patients with different hospitalization frequencies, reflecting the real-world visit patterns of cancer patients. The text length distribution shows the range of clinical note lengths per patient, highlighting the diversity in text scale within each dataset.

\section{Additional Experimental Details}
This section provides additional experimental details not fully discussed in the main text, including the process for calculating the agreement between the LLM and physicians, and some statistical results from the evaluation.

\subsection{Spearman coefficient}

To assess the agreement between a Large Language Model (LLM) and multiple human experts on the two clinical scoring tasks, we employed Spearman’s rank correlation coefficient ( $\rho$) for statistical analysis. For each evaluation metric—"Evaluation of Clinical Summary" and "Evaluation of Clinical Recommendations"—we collected the scores from the LLM (denoted as LLM Ave) and the independent scores from three experts (Exp 1, Exp 2, Exp 3). We also calculated the mean of the three expert scores (Mean). For each pair of raters, the Spearman's correlation coefficient was calculated using the following steps:

\begin{itemize}
    \item The scores from each of the two groups are converted to ranks (i.e., the original scores are replaced by their rank order within their respective groups).
    \item The Pearson correlation coefficient is then calculated between these two sets of ranks to yield the Spearman's correlation coefficient, $\rho$.
\end{itemize}

\begin{figure*}[ht]
    \centering
    \includegraphics[width=0.9\linewidth]{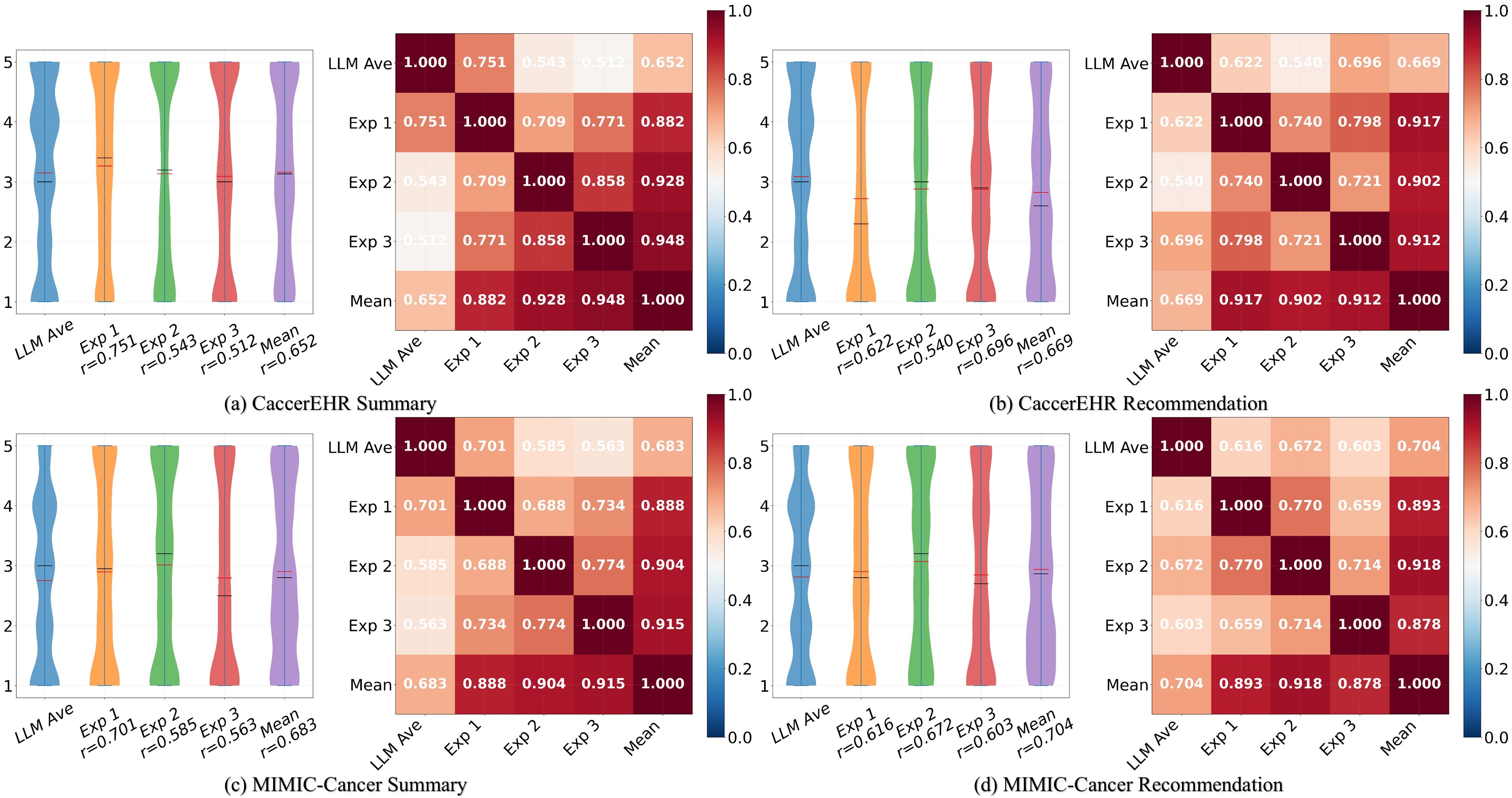}
    \caption{High Agreement Between LLM Judge and Physician Raters Validates Evaluation Methodology. The figure compares ratings from our LLM judge and three physicians on two datasets and two clinical tasks. Violin plots show similar score distributions, while heatmaps confirm high inter-rater agreement. The strong Spearman's rank correlation demonstrates that our automated evaluation is a reliable proxy for human expert judgment.}
    \label{fig:Spearman consistency.}
\end{figure*}

The resulting correlation coefficient ranges from -1 to 1, where 1 indicates a perfect positive correlation, -1 indicates a perfect negative correlation, and 0 indicates no correlation. The strength of the correlation is interpreted as follows: a coefficient greater than 0.8 is considered very strong; 0.6 to 0.8 is strong; 0.4 to 0.6 is moderate; 0.2 to 0.4 is weak; and 0 to 0.2 is very weak or negligible.The detailed formula for calculating the Spearman's coefficient is provided below:

\begin{equation}
\rho=1-\frac{6 \sum_{i=1}^{n} ( R ( X_{i} )-R ( Y_{i} ) )^{2}} {n ( n^{2}-1 )}
\end{equation}
where $n$ is the number of samples,$X_i$ and $Y_i$ are the scores of the two raters for the i-th sample,and $R(X_i)$ and $R(Y_i)$ are their respective ranks.

\subsection{Evaluation of statistical analysis}

Table \ref{tab:Statistical_Analysis} presents a detailed statistical comparison of performance for different types of large language models across two key clinical tasks. For this analysis, the evaluation results are structured into two main sections. The first section details the performance statistics for a locally deployed open-source model, Qwen-3-8B. The second section, in contrast, presents the corresponding scores from several representative closed-source models accessed via API.
This comparative analysis was conducted on two distinct medical datasets, CancerEHR and MIMIC-Cancer, and was further broken down by two independent evaluation tasks: Clinical Summary and Clinical Recommendation. To provide a comprehensive overview of the performance distribution and stability, five key descriptive statistics were calculated for each scenario: the mean, standard deviation (Std), first quartile (Q1), median, and third quartile (Q3). This granular presentation aims to thoroughly reveal the performance characteristics of each model, including their advantages, central tendencies, and the dispersion of their scores under specific clinical tasks and data environments.

\begin{table*}[t!]
    \centering
    \small
    \setlength{\tabcolsep}{2pt}
    \caption{Comparative Statistical Analysis of Performance for Different LLM Types on Clinical Tasks across the CancerEHR and MIMIC-Cancer Datasets.}
    \label{tab:Statistical_Analysis}
    \resizebox{\textwidth}{!}{
    \begin{tabular}{@{}lccccccccccccccccccccc@{}}
    \toprule
    \multirow{3}{*}{\textbf{Method}} & \multicolumn{10}{c}{\textbf{CancerEHR}} & \multicolumn{10}{c}{\textbf{MIMIC-Cancer}} \\ \cmidrule(lr){2-11} \cmidrule(lr){12-21}
     & \multicolumn{5}{c}{\textbf{$T_\textbf{CS}$}} & \multicolumn{5}{c}{\textbf{$T_\textbf{CR}$}} & \multicolumn{5}{c}{\textbf{$T_\textbf{CS}$}} & \multicolumn{5}{c}{\textbf{$T_\textbf{CR}$}} \\ \cmidrule(lr){2-6} \cmidrule(lr){7-11} \cmidrule(lr){12-16} \cmidrule(lr){17-21}
     & mean & std & Q1 & median & Q3 & mean & std & Q1 & median & Q3 & mean & std & Q1 & median & Q3 & mean & std & Q1& median & Q3 \\ \midrule
     \textit{Qwen-3-8B} \\
    \quad StandardRAG & 1.171 & 0.537 & 1 & 1 & 1 & 1.163 & 0.576 & 1 & 1 & 1 & 1.919 & 1.140 & 1 & 1 & 3 & 1.862 & 1.021 & 1 & 2 & 3 \\
    \quad BriefContext & 1.933 & 0.974 & 1 & 2 & 3 & 2.020 & 1.005 & 1 & 2 & 3 & 2.330 & 1.160 & 1 & 2 & 3 & 2.189 & 1.048 & 1 & 2 & 3 \\ 
    \quad MedRAG & 2.094 & 1.306 & 1 & 2 & 3 & 1.948 & 1.182 & 1 & 1 & 3 & 2.030 & 1.269 & 1 & 1 & 3 & 1.964 & 1.137 & 1 & 1 & 3 \\ 
    \quad KG2RAG & 2.011 & 1.159 & 1 & 2 & 3 & 2.021 & 1.191 & 1 & 1.5 & 3 & 2.193 & 1.408 & 1 & 2 & 3 & 2.226 & 1.396 & 1 & 2 & 3  \\
    \quad GNN-RAG & 1.244 & 0.575 & 1 & 1 & 1 & 1.284 & 0.615 & 1 & 1 & 1 & 1.447 & 0.835 & 1 & 1 & 2 & 1.383 & 0.752 & 1 & 1 & 2 \\
    \midrule
    \quad CliCARE & 2.446 & 1.361 & 1 & 2 & 4 & 2.385 & 1.336 & 1 & 2 & 4 & 2.046 & 1.217 & 1 & 2 & 3 & 2.015 & 1.167 & 1 & 2 & 3 \\
    \midrule
    \textit{Gemini 2.5 Pro} \\
    \quad StandardRAG & 1.804 & 1.321 & 1 & 1 & 2 & 1.781 & 1.300 & 1 & 1 & 2 & 3.863 & 1.244 & 3 & 4 & 5 & 3.832 & 1.244 & 3 & 4 & 5  \\
    \quad BriefContext & 3.933 & 1.003 & 4 & 4 & 5 & 3.924 & 0.920 & 4 & 4 & 4 & 4.442 & 0.996 & 4 & 5 & 5 & 4.376 & 1.001 & 4 & 5 & 5 \\
    \quad MedRAG & 4.526 & 0.977 & 4 & 5 & 5 & 4.543 & 1.037 & 5 & 5 & 5 & 4.477 & 1.252 & 5 & 5 & 5 & 4.492 & 1.176 & 5 & 5 & 5 \\
    \quad KG2RAG & 3.855 & 1.299 & 3 & 4 & 5 & 3.858 & 1.348 & 3 & 4 & 5 & 4.381 & 1.295 & 5 & 5 & 5 & 4.396 & 1.304 & 5 & 5 & 5 \\
    \quad GNN-RAG & 2.026 & 1.045 & 1 & 2 & 3 & 2.010 & 0.995 & 1 & 2 & 3 & 2.792 & 1.468 & 1 & 3 & 4 & 2.716 & 1.478 & 1 & 3 & 4 \\
    \midrule
    \quad CliCARE & 4.938 & 0.389 & 5 & 5 & 5 & 4.934 & 0.392 & 5 & 5 & 5 & 4.198 & 1.327 & 4 & 5 & 5 & 4.213 & 1.272 & 4 & 5 & 5 \\
    \bottomrule
    \end{tabular}
    }
\end{table*}

\section{Detailed Experimental Results on NLP Metrics}
This section presents the detailed scores of the locally deployed open-source models on the following Natural Language Processing (NLP) metrics: BLEU, ROUGE-1, ROUGE-2, and ROUGE-L. The results are summarized in the table ~\ref{tab:comparison_with_without_kg_nlp} and ~\ref{tab:ablation_nlp} below.

Specifically, Table~\ref{tab:comparison_with_without_kg_nlp} compares the performance of various models when utilizing a standard RAG setup versus our proposed CliCARE framework. Furthermore, Table ~\ref{tab:ablation_nlp} details the results of an ablation study on the key components of the CliCARE framework, quantifying the contribution of each module to the overall performance.

It is important to note that for an open-ended medical generation task like ours, standard NLP metrics such as BLEU and ROUGE have inherent limitations. These metrics primarily measure the lexical similarity between the generated text and the reference answers. Consequently, while they can provide a general indication of fluency and content overlap, they cannot fully capture the clinical authenticity or factual accuracy of the generated responses.

\begin{table*}[t]
    \centering
    \small 
    \setlength{\tabcolsep}{3pt} 
    \caption{
        Model performance with standard RAG versus the CliCARE framework. 
        The table shows scores for BLEU (B), ROUGE-1 (R1), ROUGE-2 (R2), and ROUGE-L (RL). 
        Applying CliCARE provides a substantial performance uplift, especially on the complex CancerEHR dataset.
    }
    \label{tab:comparison_with_without_kg_nlp} 
    \resizebox{\textwidth}{!}{
    \begin{tabular}{@{}lcccccccccccccccc@{}}
    \toprule
    \multirow{3}{*}{\textbf{Method}} & \multicolumn{8}{c}{\textbf{Standard RAG}} & \multicolumn{8}{c}{\textbf{CliCARE}} \\ 
    \cmidrule(lr){2-9} \cmidrule(lr){10-17} 
     & \multicolumn{4}{c}{\textbf{CancerEHR}} & \multicolumn{4}{c}{\textbf{MIMIC-Cancer}} & \multicolumn{4}{c}{\textbf{CancerEHR}} & \multicolumn{4}{c}{\textbf{MIMIC-Cancer}} \\ 
    \cmidrule(lr){2-5} \cmidrule(lr){6-9} \cmidrule(lr){10-13} \cmidrule(lr){14-17}
     & B & R1 & R2 & RL & B & R1 & R2 & RL & B & R1 & R2 & RL & B & R1 & R2 & RL \\ \midrule
    Mistral-v0.1-7B & \textbf{42.46} & \textbf{54.52} & \textbf{28.52} & \textbf{44.69} & \textbf{68.71} & \textbf{54.80} & \textbf{31.90} & \textbf{40.18} & 38.42 & 50.34 & 23.25 & 34.55 & \textbf{68.04} & \textbf{53.20} & \textbf{30.53} & \textbf{38.99} \\
    Mistral-Instruct-v0.1-7B & 41.83 & 53.45 & 27.86 & 44.58 & 67.80 & 54.11 & 31.25 & 39.16 & 37.90 & 49.79 & 22.86 & 34.33 & 67.54 & 52.34 & 29.95 & 38.67 \\
    Biomistral-7B & 41.20 & 53.14 & 27.58 & 43.87 & 66.81 & 53.42 & 31.05 & 39.28 & 29.98 & 41.07 & 18.42 & 28.07 & 66.50 & 51.62 & 29.32 & 38.28 \\
    Qwen-3-8B & 22.81 & 39.59 & 18.24 & 26.49 & 67.73 & 53.07 & 29.94 & 38.38 & \textbf{39.88} & \textbf{53.16} & \textbf{24.55} & \textbf{35.23} & 66.90 & 52.09 & 29.02 & 37.58 \\
    \midrule
    \end{tabular}
    }
\end{table*}

\begin{table*}[h]
    \centering
    \small 
    \setlength{\tabcolsep}{4pt} 
    \caption{
        Ablation study on CliCARE framework components using NLP metrics. 
    }
    \label{tab:ablation_nlp} 
    \begin{tabular}{@{}lcccccccc@{}}
        \toprule
        \multirow{2}{*}{\textbf{Method}} & \multicolumn{4}{c}{\textbf{CancerCEHR}} & \multicolumn{4}{c}{\textbf{MIMIC-Cancer}} \\
        \cmidrule(lr){2-5} \cmidrule(lr){6-9}
        & B & R1 & R2 & RL & B & R1 & R2 & RL \\
        \midrule
        \textbf{CliCARE (Qwen-3-8B)} & 39.88 & 53.16 & 24.55 & 35.23 & \textbf{66.90} & \textbf{52.09} & \textbf{29.02} & 37.58 \\
        \quad w/o Alignment Expansion & 42.34 & 58.07 & 31.08 & 42.40 & 62.33 & 50.19 & 28.04 & \textbf{44.57} \\
        \quad w/o LLM-based Reranking & \textbf{43.58} & \textbf{58.71} & \textbf{31.45} & \textbf{42.65} & 61.59 & 49.55 & 27.54 & 44.06 \\
        \quad w/o TKG-based Compression & 22.81 & 39.59 & 18.24 & 26.49 & 67.73 & 53.07 & 29.94 & 38.38 \\
        \midrule
    \end{tabular}
\end{table*}

\section{Effects of Context Length on Performance}
This section elaborates on the sensitivity analysis conducted to evaluate the effect of different context lengths on model performance. When processing long-form medical documents such as EHRs, the model's context processing capability is crucial, as it directly determines the accuracy and comprehensiveness of its information integration, clinical summarization, and decision support. To quantify this effect, we selected the Qwen3-8B model and conducted experiments on two medical datasets: CancerEHR and MIMIC-Cancer. We systematically configured and tested three different context lengths: 2k, 8k, and 20k tokens. By comparing the performance under these configurations, we aim to reveal the relationship between model performance and context length.

\begin{figure}
    \centering
    \includegraphics[width=0.5\textwidth]{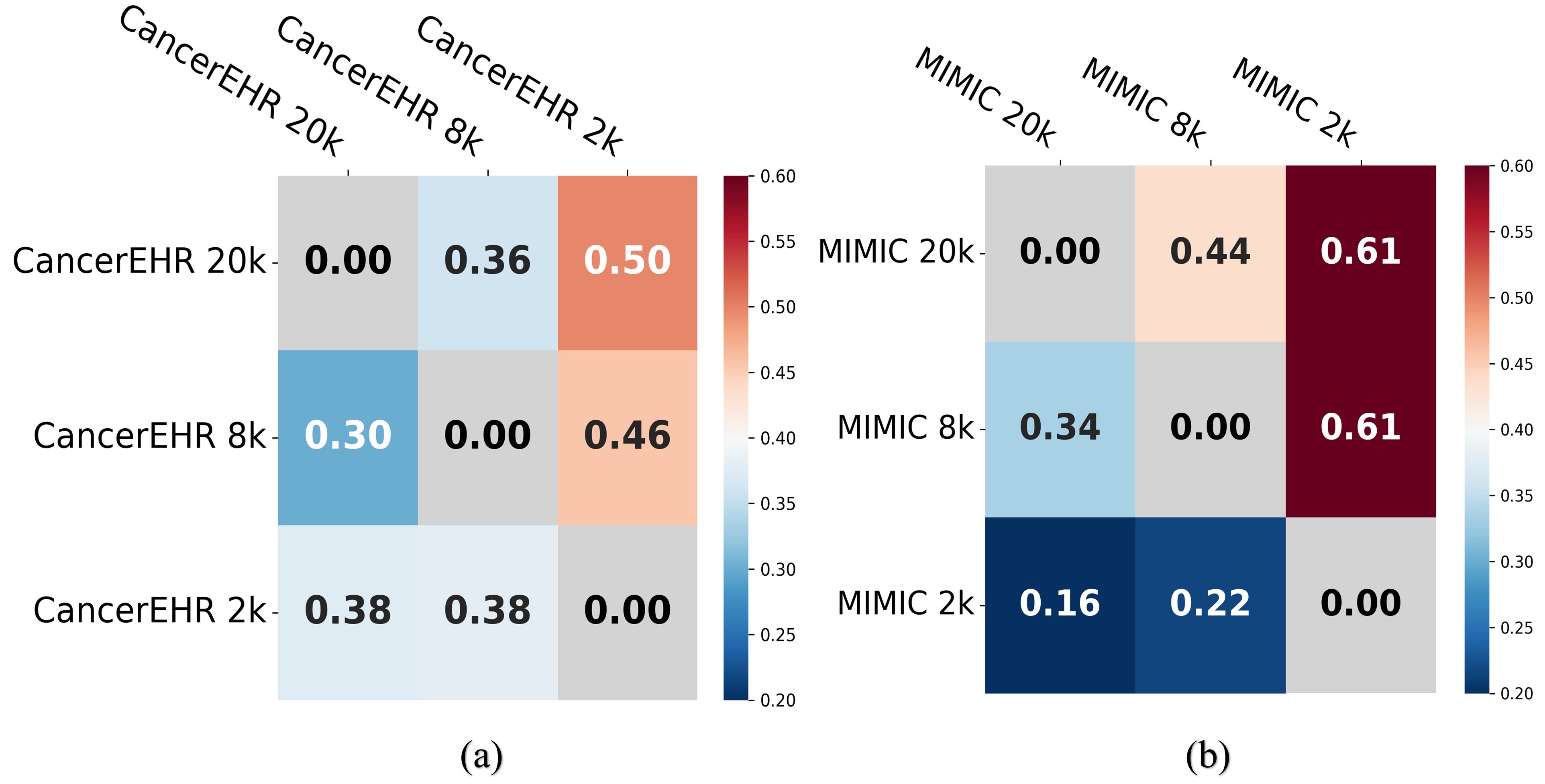}
    \caption{Ablation study results for the model at different context lengths (2k, 8k, 20k). (a) Pairwise comparison matrix on the CancerEHR dataset. (b) Pairwise comparison matrix on the MIMIC-Cancer dataset.}
    \label{fig:Contextual_Confusion_Matrix}
\end{figure}

As shown in Figure \ref{fig:Contextual_Confusion_Matrix}, these two matrices illustrate the results of the performance ablation study under different context length settings. We compare the performance of three context lengths—2k, 8k, and 20k tokens—on two datasets: CancerEHR (Figure a) and MIMIC-Cancer (Figure b). The values in the matrix represent the win rate of the row model over the column model in pairwise comparisons. Across both datasets, we found that increasing the model's context length generally improves its performance. On the CancerEHR dataset (Figure a), the 20k model has a win rate of 0.50 against the 2k model, and the 8k model has a win rate of 0.46 against the 2k model. The competition between the 20k and 8k models is closer, with their respective win rates being 0.36 and 0.30, indicating a slight advantage for the 20k version. Similarly, on the MIMIC-Cancer dataset, the advantage of the 20k model over the 8k model is also pronounced, with a win rate of 0.44, while the win rate of the 8k model against the 20k model is only 0.34. These results demonstrate the importance of a long context for improving model performance when processing complex Electronic Health Record (EHR) data, suggesting that the model can more effectively utilize the extended contextual information to make higher-quality judgments, leading to more significant performance gains.

\section{Implementation Details of the Alignment Method}
This section provides the detailed pseudocode for our proposed Guideline and Patient Data Alignment method. As outlined in ~\ref{alg:Alignment_Algorithm}, the algorithm consists of five main stages. 

First, it employs a Large Language Model (LLM) to construct a knowledge graph (KG) from clinical guidelines and a temporal knowledge graph (TKG) from patient records (Step 1). Following this, an initial set of alignment candidates is generated using BERT-based semantic similarity (Step 2) and subsequently reranked by an LLM to create a high-quality seed set ($A_{\text{seed}}$) based on clinical plausibility (Step 3). The core of the method is the iterative BootEA process, which expands these seed alignments by calculating a weighted score derived from both semantic and neighborhood similarities for all unaligned pairs (Step 4). The process concludes by merging the KG and TKG based on the final alignment set to produce a single, unified graph (Step 5).

\begin{algorithm*}[btp]
\caption{Alignment Algorithm}
\label{alg:Alignment_Algorithm}
\begin{algorithmic}[1]

\Require
    Clinical Guidelines $D_g$;
    Patient Clinical Data $D_p$;
    Pre-trained BERT model $f_{\text{BERT}}$;
    Large Language Model $f_{\text{LLM}}$;
    Bootstrapping iterations $I$;
    Confidence threshold $\theta$;
    Weighting factor $\alpha$ for scoring.
\Ensure
    An aligned Knowledge Graph $G_{\text{aligned}}$.

\Statex
\Statex \textbf{Step 1: Knowledge Graph Extraction}
\State Utilize $f_{\text{LLM}}$ to extract a static $KG$ from $D_g$ and a Temporal $TKG$ from $D_p$.

\Statex
\Statex \textbf{Step 2: Initial Semantic Alignment with BERT}
\State Initialize candidate set $A_{\text{initial}} \leftarrow \emptyset$.
\For{each entity $e_p \in TKG$}
    \For{each entity $e_g \in KG$}
        \State $sim \leftarrow \text{cosine\_similarity}(f_{\text{BERT}}(\text{desc}(e_p)), f_{\text{BERT}}(\text{desc}(e_g)))$.
        \State Add $(e_p, e_g, sim)$ to $A_{\text{initial}}$.
    \EndFor
\EndFor

\Statex
\Statex \textbf{Step 3: LLM Reranking}
\State $A_{\text{reranked}} \leftarrow f_{\text{LLM}}(\text{"Rerank candidates by clinical plausibility"}, A_{\text{initial}})$.
\State $A_{\text{seed}} \leftarrow \text{FilterHighConfidencePairs}(A_{\text{reranked}})$.

\Statex
\Statex \textbf{Step 4: Iterative Bootstrapping Alignment}
\State $A_{\text{final}} \leftarrow A_{\text{seed}}$.
\State Let $U$ be the set of unaligned entities.
\For{$i = 1 \to I$}
    \Statex \textit{Initialize a set for newly found alignments in this iteration:}
    \State $A_{new} \leftarrow \emptyset$.
    \For{each unaligned pair $(e_p, e_g) \in U \times U$}
        \Statex \textit{Calculate semantic similarity:}
        \State $S_{sem} \leftarrow \text{cosine\_similarity}(f_{\text{BERT}}(e_p), f_{\text{BERT}}(e_g))$.
        \Statex \textit{Calculate neighborhood similarity based on already aligned neighbors:}
        \State $S_{hood} \leftarrow \text{CalculateNeighborhoodScore}((e_p, e_g), A_{\text{final}})$.
        \Statex \textit{Compute the final weighted score:}
        \State $S_{weighted} \leftarrow \alpha \cdot S_{sem} + (1-\alpha) \cdot S_{hood}$.
        
        \If{$S_{weighted} > \theta$}
            \State $A_{new} \leftarrow A_{new} \cup \{(e_p, e_g)\}$.
        \EndIf
    \EndFor
    
    \If{$A_{new}$ is empty}
        \Statex \textit{Exit loop if no new alignments are found.}
        \State \textbf{break}
    \EndIf

    \Statex \textit{Add newly found alignments to the final set:}
    \State $A_{\text{final}} \leftarrow A_{\text{final}} \cup A_{new}$.
    \Statex \textit{Update the set of unaligned entities:}
    \State $U \leftarrow U \setminus \text{entities in } A_{new}$.
\EndFor

\Statex
\Statex \textbf{Step 5: Construct Aligned Graph}
\State $G_{\text{aligned}} \leftarrow \text{MergeGraphs}(KG, TKG, A_{\text{final}})$.

\Statex
\State \Return $G_{\text{aligned}}$.
\end{algorithmic}
\end{algorithm*}

\section{PROMPT OF Answer Label GENERATION}
\begin{figure*}
    \centering
    \includegraphics[width=0.9\linewidth]{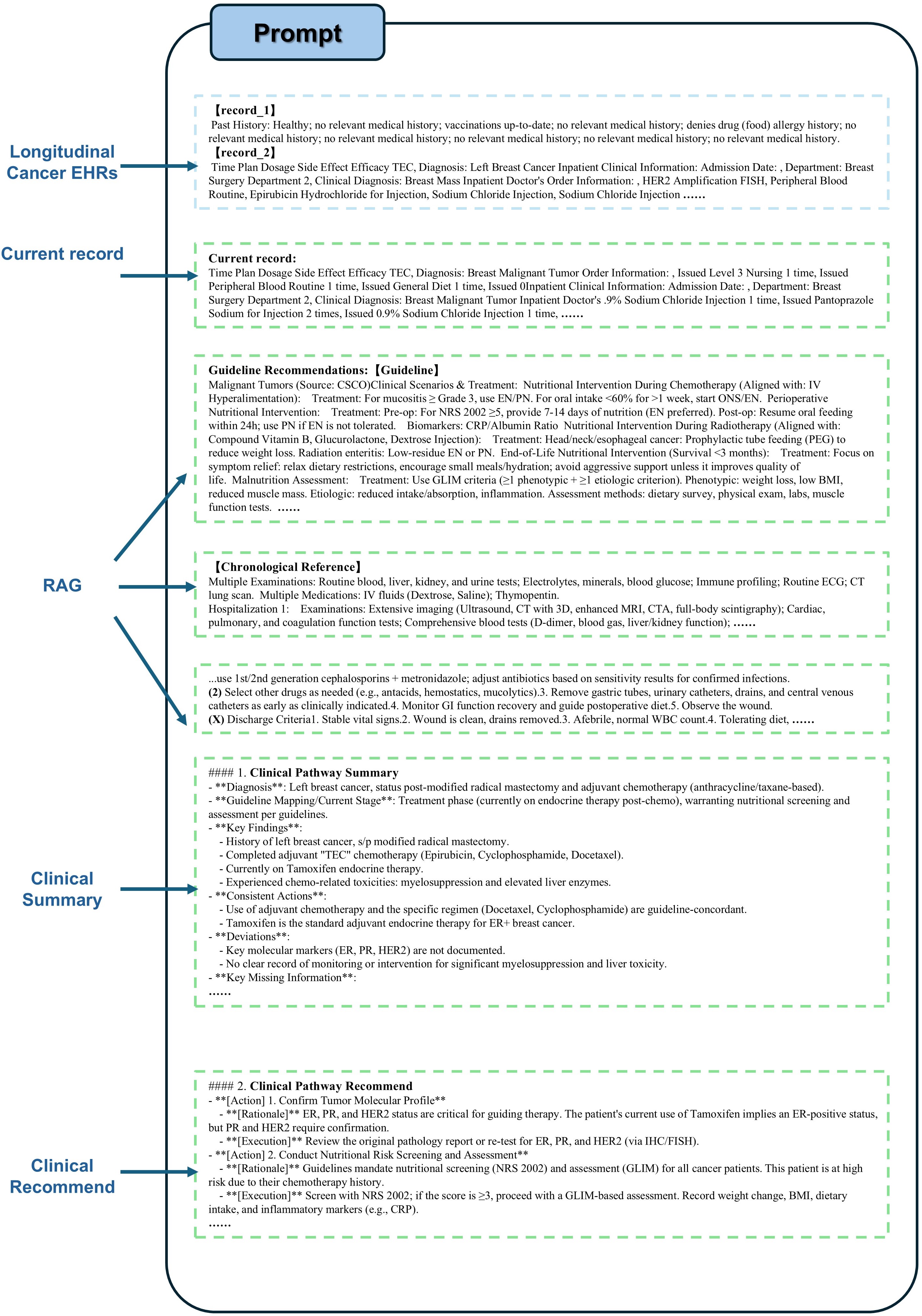}
    \caption{A Sample Prompt for the CliCARE Method Highlighting Its Key Components.}
    \label{fig:prompt_case}
\end{figure*}

As shown in Figure \ref{fig:prompt_case}, this is a sample prompt for the CliCARE method. The prompt is structured into five primary components: Longitudinal Cancer EHRs, Current record, Retrieval-Augmented Generation (RAG) content, a section for the Clinical Summary, and a section for the Clinical Recommendation. These elements combine to form a single, comprehensive prompt that guides the model's response generation process.

\end{document}